\pdfoutput=1

\documentclass[11pt,a4paper]{article}
\usepackage{times}
\usepackage{latexsym}
  \usepackage{hyperref}
\hypersetup{
  pdfinfo={
    Title={SIRIUS-LTG-UiO at SemEval-2018 Task 7:\\Convolutional Neural Networks with Shortest Dependency Paths for Semantic Relation Extraction and Classification in Scientific Papers},
    Author={YFarhad Nooralahzadeh, Lilja Øvrelid, Jan Tore Lønning}
  }
}
\usepackage{pdfpages}
\usepackage[utf8]{inputenc}

\usepackage[nohyperref]{naaclhlt2018}
\usepackage{url}
\usepackage{enumitem}
\usepackage{caption}
\usepackage{graphicx}
\usepackage{lipsum}
\usepackage{multirow}
\usepackage{caption}
\usepackage{graphicx}
\usepackage{tabularx}
\usepackage{array}
\usepackage{booktabs}
\usepackage{nth}
\newcommand{\head}[1]{\textnormal{\textbf{#1}}}
\usepackage{amsmath}
\usepackage{textcomp}
\newcolumntype{P}[1]{>{\centering\arraybackslash}p{#1}}

\makeatletter
\newcounter{subsubparagraph}[subparagraph]
\renewcommand\thesubsubparagraph{%
  \thesubparagraph.\@arabic\c@subsubparagraph}
\newcommand\subsubparagraph{%
  \@startsection{subsubparagraph}    % counter
    {6}                              % level
    {\parindent}                     % indent
    {.3ex \@plus .1ex \@minus .1ex} % beforeskip
    {-.6em}                           % afterskip
    {\normalfont\normalsize\bfseries}}
\newcommand\l@subsubparagraph{\@dottedtocline{6}{10em}{5em}}
\newcommand{\subsubparagraphmark}[1]{}
\makeatother

\newcounter{eqnnosave}          % used in trick with equation number

\newenvironment{exlist}{%         % define "example" environment
   \begin{enumerate}%
   \setcounter{enumi}{\arabic{eqnnosave}}%   % restores previous value
}%
{\end{enumerate}%
\setcounter{eqnnosave}{\arabic{enumi}}%
}

\aclfinalcopy % Uncomment this line for all SemEval submissions

\title{ SIRIUS-LTG-UiO at SemEval-2018 Task 7:\\Convolutional Neural Networks with Shortest Dependency Paths for Semantic Relation Extraction and Classification in Scientific Papers}
\author{Farhad Nooralahzadeh, Lilja Øvrelid, Jan Tore Lønning \\
  Department of Informatics\\ University of Oslo, Norway\\
  {\tt $\{$farhadno,liljao,jtl$\}$@ifi.uio.no} \\}

\date{}

\begin{document}
\maketitle
\begin{abstract}
This article presents the SIRIUS-LTG-UiO system for the SemEval 2018 Task 7 on Semantic Relation Extraction and Classification in Scientific Papers. First we extract the shortest dependency path (sdp) between two entities, then we introduce a convolutional neural network (CNN) which takes the shortest dependency path embeddings as input and performs relation classification with differing objectives for each subtask of the shared task. This approach achieved overall F1 scores of 76.7 and 83.2 for relation classification on clean and noisy data, respectively. Furthermore, for combined relation extraction and classification on clean data, it obtained F1 scores of 37.4 and 33.6 for each phase. 
Our system ranks 3rd in all three sub-tasks of the shared task.
\end{abstract}

\section{Introduction}
Relation extraction and classification can be defined as
follows: given a sentence where entities are manually annotated, we aim to identify the pairs of entities that are instances of the semantic relations of interest and classify them based on a pre-defined set of relation types.  A range of different approaches have been applied to solve this task in previous work. Conventional classification approaches have made use of contextual, lexical and syntactic features combined with richer linguistic and background knowledge such as WordNet and
FrameNet \cite{Hendrickx:2010:STM:1859664.1859670,Rink:2010:UCS:1859664.1859721}.

Recently, the re-emergence of deep neural networks provides a way to develop highly automatic features and representations to handle complex interpretation tasks. These approaches have yielded impressive results for many different NLP tasks. The use of deep neural networks for relation classification has been investigated in several recent studies \cite{Socher:2012:SCT:2390948.2391084,P16-1200,DBLP:conf/acl/ZhouSTQLHX16}.
Convolutional neural networks (CNNs) 
%as a type of network architecture for deep learning is comprised of one or more convolutional layers followed by a pooling operation and then includes one or more fully connected layers at the end. CNNs 
have been effectively applied to extract lexical and sentence level features for relation classification \cite{DBLP:journals/corr/ZhangW15a,DBLP:journals/corr/LeeDS17,W15-1506}. However, these works consider whole sentences or the context between two target entities as input for the CNN. Such representations suffer from irrelevant sub-sequences or clauses when target entities occur far from each other or there are other target entities in the same sentence. To avoid negative effects from irrelevant chunks or clauses and capture the relation between two entities, \citet{DBLP:journals/corr/XuFHZ15,DBLP:journals/corr/LiuWLJZW15} and \citet{DBLP:journals/corr/XuMLCPJ15} employ a CNN to learn more robust and effective relation representations from the shortest dependency path (sdp) between two entities. The sdp between two entities in the dependency graph captures a condensed representation of the information required  to assert a relationship between two entities \cite{Bunescu:2005:SPD:1220575.1220666}. In this work, we continue this line of work and present a system based on a CNN architecture over shortest dependency paths combined with domain-specific word embeddings to extract and classify semantic relations in scientific papers.

\section{System description}
In this section, we describe the various components of our
system.
\paragraph{Text pre-processing.} For each relation instance in the training data set, we assign a sentence that contains the participant entities. Sentence and token boundaries are detected using the Stanford CoreNLP tool \cite{manning-EtAl:2014:P14-5}. Since most of the entities are multi-word units, in order to obtain a precise dependency path between entities, we replace the entities with their codes. The example sentence in (\ref{ex:sent}) below is thus transformed to (\ref{ex:ent}).
\begin{exlist}
    \item\label{ex:sent} Syntax-based statistical machine translation (MT) aims at applying statistical models to structured data .
    \item\label{ex:ent} P05-1067.1 aims at applying P05-1067.2 to P05-1067.3 .
\end{exlist}
 Given an encoded sentence, we find the sdp connecting two target entities for each relation instance using a syntactic parser, see below.

%Added by lilja 
For syntactic parsing we employ the parser described in \newcite{Boh:Niv:12}, a transition-based parser which performs joint PoS-tagging and parsing. We train the parser on the standard training sections 02-21 of the Wall Street Journal (WSJ) portion of the Penn Treebank \cite{Mar:San:Mar:93}. The constituency-based treebank is converted to dependencies using two different conversion tools: (i) the pennconverter
software\footnote{\url{http://nlp.cs.lth.se/software/treebank-converter/}} \cite{Joh:Nug:07}, which produces the so-called CoNLL-style dependencies employed in the CoNLL08 shared task on dependency parsing \cite{Sur:Joh:Mey:08}\footnote{The pennconverter tool is run using the \texttt{rightBranching=false} flag.}, and (ii) the Stanford parser using the option to produce basic Stanford dependencies \cite{Mar:Doz:Sil:14}\footnote{The Stanford parser is run using the \texttt{-basic} flag to produce the basic version of Stanford dependencies.}.
The parser achieves a labeled accuracy score of 91.23 when trained on the CoNLL08 representation and 91.31 for the Stanford basic model, when evaluated against the standard evaluation set (section 23) of the WSJ.  We also experimented with the pre-trained parsing model for English included in the Stanford CoreNLP toolkit \cite{manning-EtAl:2014:P14-5}, which outputs Universal Dependencies. However, it was clearly outperformed by our version of the \newcite{Boh:Niv:12} parser in the initial development experiments.

Based on the dependency graphs output by the parser, we extract the shortest dependency path connecting two entities. The path records the direction of arc traversal
using left and right arrows (i.e. $\leftarrow$ and $\rightarrow$) as well as the dependency relation of the traversed arcs and the predicates involved, following \newcite{DBLP:journals/corr/XuFHZ15}. The entity codes in the final sdp are replaced with the corresponding word tokens at the end of the pre-processing step. For the sentence in (\ref{ex:sent}) and the two entities \textit{statistical models} and \textit{structured data} we thus extract the path in (\ref{ex:path}) below.
\begin{exlist}
    \item\label{ex:path}\texttt{statistical models $\leftarrow$ OBJ $\leftarrow$ applying $\rightarrow$ DIR $\rightarrow$ to $\rightarrow$ PMOD $\rightarrow$ structured data}
\end{exlist}

\paragraph{Label encoding.}\label{encoding} The classification sub-tasks contain five asymmetric relations (USAGE, RESULT, MODEL-FEATURE, PART\_WHOLE, TOPIC)  and one symmetric relation (COMPARE). The relation instance along with its directionality are provided in both the training and the test data sets. For these sub-tasks we therefore use the same labels in our system. 
For sub-task 2 which combines the extraction and classification tasks, however, we construct an extra set of relation types. First, we collect every pair of entities within a single sentence that are not included in the annotated relation set. To minimize the noise, we retain only the entity pairs which are not further away than $6$ tokens. From these entity pairs we generate negative instances with the NONE class and extract the corresponding sdp. Second, to preserve the directionality in the asymmetric relations, we add the $\neg$ symbol to the instances with reverse directionality (e.g., USAGE(e1,e2,REVERSE) becomes  $\neg$USAGE(e1,e2)). The final label set for sub-task 2 thus consists of 12 relations.  

\paragraph{Word embeddings.} In our system, two different sets of pre-trained word embeddings are used for initialization. One is the 300-d pre-trained embeddings provided by the NLPL repository \footnote{\url{http://vectors.nlpl.eu/repository/}}\cite{Velldal}, trained on English Wikipedia data with word2vec \cite{Mikolov2013a}, here  dubbed wiki-w2v. In addition, we train a second set of domain-specific embeddings on the ACL Anthology corpus. We obtain the XML versions of 22,878 articles from ACL Anthology \footnote{\url{https://acl-arc.comp.nus.edu.sg/}}. After extracting the raw texts, for training of the 300-d word embeddings (acl-w2v), we exploit the available word2vec \cite{Mikolov2013a} implementation \emph{gensim} \cite{gensim} for training.  
%Beyond the word embedding models, our system does not use any other external resources.
\paragraph{Classification Model}
\begin{figure*}[t]
\centering
\includegraphics[width=.95\textwidth]{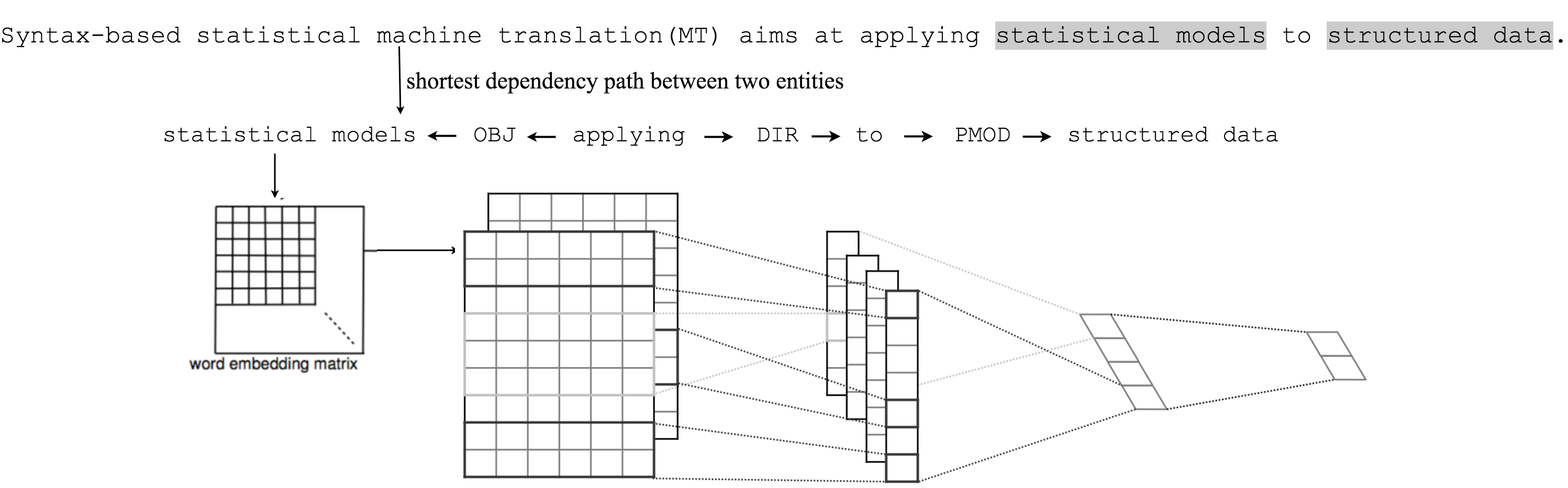}
\caption{Model architecture with two channels for an example shortest dependency path (CNN model from \newcite{DBLP:journals/corr/Kim14f}).}
\label{fig:cnn}
\end{figure*} 
Our system is based on a Convolutional Neural Network (CNN) architecture similar to the one used for sentence classification in \newcite{DBLP:journals/corr/Kim14f}. Figure \ref{fig:cnn} provides an overview of the proposed model. It consists of 4 main layers as follows:
\subsubparagraph{Look-up Table and Embedding layer:} In the first step, the model takes a dependency path, as in (\ref{ex:path}) as input and transforms it into a matrix representation by looking up the pre-trained word embeddings.
\subsubparagraph{Convolutional Layer:}
The next layer performs convolutions with the ReLU activation to the embedding layer using multiple filter sizes (\emph{filter\_sizes} $\in [3,4,5]$) and extracts feature maps over the tokens.
\subsubparagraph{Max pooling Layer:}
By applying the \emph{max} operator, the most effective local features are generated from each feature map.
\subsubparagraph{Fully connected Layer:}
Finally, the higher level syntactic features are fed to a fully connected \emph{softmax} layer which outputs the probability distribution over each relation.

\section{Experiments}

\paragraph{Dataset} For each sub-task, the training data includes abstracts of papers from the ACL Anthology corpus with pre-annotated entities. For sub-task 1.1 and 2, the training datasets are the same. It contains entities that are manually annotated and they represent domain concepts specific to Natural Language Processing (NLP). In sub-task 1.2 the entities are automatically assigned and therefore contain a fair amount of noise (verbs, irrelevant words). The terms include high-level terms (e.g. "algorithm", "paper", "method") and are not always full NPs \cite{SemEval2018Task7}.
Since the related entity pairs and the relation types are provided for the full dataset, we extend the dataset for sub-task 1.1 and 2 by extracting the related entities and their corresponding sdp from the sub-task 1.2 dataset. In order to train a model for sub-task 2, we also augment the dataset by extracting NONE relation instances (see Section \ref{encoding}), extracted from the corresponding dataset. Table \ref{dataset} shows the number of instances for each relation class. As we can see, the class distribution is clearly unbalanced.
\begin{table}[t]
  \centering
    \scalebox{.85}{
  \begin{tabular}{*{6}{lrrrrr}}
    \toprule
    &\multicolumn{2}{c}{\head{Subtask}} & \multicolumn{2}{c}{\head{Reverse}} & \\
    \cmidrule(ll){2-3}
    \cmidrule(ll){4-5}
    \head{Relation} &1.1 \& 2 & 1.2& False & True & \head{Total} \\
    \midrule
    {\small USAGE} &483 & 464 &615 &332 &947    \\
        {\small MODEL-FEATURE} & 326& 172&346 & 152    & 498 \\  
        {\small RESULT}&72&121 & 135 & 58 & 193 \\
         {\small TOPIC} &18&240& 235 & 23 &  258\\
         {\small PART\_WHOLE}&233& 192& 273 & 152 & 425 \\
          {\small COMPARE} &95&41& 136 & - &136 \\
          {\small NONE} & 2315&-&2315& - & 2315 \\
    \bottomrule
  \end{tabular}}
   \caption{Number of instances for each relation in the final dataset.}
   \label{dataset}
\end{table}

\begin{table*}[t]
  \centering
  \scalebox{.87}{
  \begin{tabular}{*{5}{lp{4cm}p{4cm}rr}}
  \toprule
    &&&\multicolumn{2}{c}{\head{F1}}\\
    \cmidrule(lr){4-5}
    \head{Sub-task} & \head{Model} & \head{Representation} & Ext. & Class. \\
    \midrule
     1.1 &\multirow{ 2}{*}{cnn.multi.acl-w2v.rand}& \multirow{ 2}{*}{Stanford Basic} & - &74.16  \\
     1.2 & & &- & 77.70  \\
     2 & cnn.acl-w2v & CoNLL08 &74.26& 60.31\\
    \bottomrule
  \end{tabular}
  }
   \caption{F1 (macro-average) scores for selected configurations during training.}
   \label{result}
\end{table*}
\paragraph{Model settings}
We keep  the  value  of hyperparameters equal to the ones that are reported in the original work \cite{DBLP:journals/corr/Kim14f}, i.e., $128$ filters for each window size, a dropout rate of $\rho=0.5$ and $l_2$ regularization of $3$.
To deal with the effects of class imbalance,  we weight the cost by the ratio of class instances, thus each observation receives a weight, depending on the class it belongs to. The effect of the minority class observations is thereby increased simply by a higher weight of these instances and is decreased for majority class observations. Furthermore, to guarantee that each fold in $n$-fold cross validation will have the proportion of same classes during training, evaluation and test, we apply the stratification technique proposed by \citet{Sechidis:2011:SMD:2034161.2034172}.
We use the validation set to detect when overfitting starts during
the training of our model; using \emph{early stopping}, training is then stopped before convergence to avoid  overfitting \cite{Prechelt:1998:ES:645754.668392}. The official evaluation metric is the macro-averaged F1-score, therefore we implement early-stopping (\emph{patience}= $20$) based on macro-F1 score in the development set.
\paragraph{Model variants}
We run experiments with several variants of the model as follows:  \texttt{cnn.rand:} A baseline model, where all elements in the embedding layer are randomly initialized and updated in the training process. \texttt{cnn.wiki-w2v:} The embedding layer is initialized with the  pre-trained Wikipeida word embeddings and fine-tuned for the target task. \texttt{cnn.acl-w2v}: The embedding layer is initialized with the  pre-trained ACL Anthology word embeddings and fine-tuned for the target task. \texttt{cnn.multi.rand:} There are two embedding layers as a 'channel' in the CNN architecture. Both channels are initialized randomly and only one of them is updated during training while the other remains static. \texttt{cnn.multi.wiki-w2v:} Same as before, but the channels are initialized with Wikipedia embedding vectors. \texttt{cnn.multi.acl-w2v:} The two channels are initialized with ACL embedding vectors.
\texttt{cnn.multi.wiki-w2v.rand:} First channel is initialized with Wikipedia embeddings in static mode and the second initialized randomly with a non-static mode. 
\texttt{cnn.multi.acl-w2v.rand:} Same  as  previous  setting, but  the  first channel makes use of ACL embeddings. 

\paragraph{Results}
During development, we investigate the performance of different configurations; different dependency representations (CoNLL08 and Stanford basic) and  model variants (see above); by running $5$-fold cross validation (i.e. $3$ folds for training, $1$ fold for evaluation and $1$ fold for test). The experiments show that, the multi-channel mode performs better only in the classification sub-tasks compared to the single channel setting. The results suggest that having a significant amount of instances per relation assists the model to classify better. The use of the pre-trained embeddings helps the model in class assignment. Particularly, the domain-specific embeddings (i.e. acl-w2v) provide higher performance gains when used in the model. Table \ref{result} presents the F1-score of the best performing model for each sub-task via $5$-fold cross validation on the training data. In the evaluation period, we re-run $5$-fold cross validation using selected model for each sub-task. However, in this setting we use 4 folds as training and 1 fold as development set, and we apply the output model to the evaluation dataset.
We select the \nth{1} and \nth{2} best performing models on the development datasets as well as the majority vote (mv) of 5 models for the final submission. The final results are shown in Table \ref{semeval}.
\begin{table}[h]
\centering
\scalebox{.8}{
\begin{tabular}{*{7}{l}}
\toprule
&  \multicolumn{2}{c}{\head{\nth{1}}} & \multicolumn{2}{c}{\head{\nth{2}}} 
      & \multicolumn{2}{c}{\head{\emph{mv}}} \\
\cmidrule(r){2-3}
\cmidrule(r){4-5}
\cmidrule(r){6-7}
\head{Sub-task} & Ext.& Class. &Ext.& Class.&Ext.& Class. \\
\midrule
1.1 &- &72.1 &-& 74.7 &-&{\bf76.7} \\
1.2 &  - & {\bf 83.2} &-& 82.9 &-& 80.1  \\
2 & {\bf 37.4} & {\bf 33.6} & 36.5& 28.8 & 35.6& 28.3 \\
\bottomrule
\end{tabular}
}
\caption{Official evaluation results of the submitted runs on the test set.}
\label{semeval}
\end{table}
\section{Conclusion}
We present a CNN model over shortest dependency paths between entity pairs for relation extraction and classification. We examine various architectures for the proposed model. The experiments demonstrate the effectiveness of domain-specific word embeddings for all sub-tasks as well as sensitivity to the specific dependency representation employed in the input layer. Our future work includes: 1) to perform error analysis for the different sub-tasks, and 2) to investigate the effects of different dependency representations in relation extraction and classification.

%\section{Acknowledgements}
%We thank the shared task organizers for their efforts and the anonymous reviewers for their helpful comments.

\bibliography{semeval2018}
\bibliographystyle{acl_natbib}

\appendix

\end{document}